\documentclass[letterpaper]{article} 
\usepackage{aaai24}  
\usepackage{times}  
\usepackage{helvet}  
\usepackage{courier}  
\usepackage[hyphens]{url}  
\usepackage{graphicx} 
\usepackage{natbib}  
\usepackage{caption} 
\usepackage{algorithm}
\usepackage{algorithmic}
\usepackage{adjustbox}
\usepackage{amsmath}
\usepackage{amsfonts}
\usepackage{tabularx, booktabs}
\usepackage[inline]{enumitem}
\usepackage{subcaption}
\usepackage{colortbl}
\newsavebox{\mybox}
\definecolor{myblue}{rgb}{0.8,0.85,1} 
\usepackage{newfloat}
\usepackage{listings}
\DeclareCaptionStyle{ruled}{labelfont=normalfont,labelsep=colon,strut=off} 
\floatstyle{ruled}
\newfloat{listing}{tb}{lst}{}
\floatname{listing}{Listing}
\title{Market-GAN: Adding Control to Financial Market Data Generation with Semantic Context}
\author{
    Haochong Xia,
    Shuo Sun,
    Xinrun Wang\thanks{Corresponding authors},
    Bo An\textsuperscript{*}
}
\affiliations{
    Nanyang Technological University, Singapore\\
    \{haochong001, shuo003\}@e.ntu.edu.sg,
    \{xinrun.wang, boan\}@ntu.edu.sg
}

\usepackage{bibentry}

\begin{document}

\maketitle

\begin{abstract}
Financial simulators play an important role in enhancing forecasting accuracy, managing risks, and fostering strategic financial decision-making. Despite the development of financial market simulation methodologies, existing frameworks often struggle with adapting to specialized simulation context. We pinpoint the challenges as i) current financial datasets do not contain context labels; ii) current techniques are not designed to generate financial data with context as control, which demands greater precision compared to other modalities; iii) the inherent difficulties in generating context-aligned, high-fidelity data given the non-stationary, noisy nature of financial data. To address these challenges, our contributions are: i) we proposed the Contextual Market Dataset with market dynamics, stock ticker, and history state as context, leveraging a market dynamics modeling method that combines linear regression and clustering to extract market dynamics; ii) we present Market-GAN, a novel architecture incorporating a Generative Adversarial Networks (GAN) for the controllable generation with context, an autoencoder for learning low-dimension features, and supervisors for knowledge transfer; iii) we introduce a two-stage training scheme to ensure that Market-GAN captures the intrinsic market distribution with multiple objectives. In the pertaining stage, with the use of the autoencoder and supervisors, we prepare the generator with a better initialization for the adversarial training stage. We propose a set of holistic evaluation metrics that consider alignment, fidelity, data usability on downstream tasks, and market facts. We evaluate Market-GAN with the Dow Jones Industrial Average data from 2000 to 2023 and showcase superior performance in comparison to 4 state-of-the-art time-series generative models.

\end{abstract}

\section{Introduction}
\label{introduction}
Financial simulations play a pivotal role in navigating the complexities of the economic landscape, enabling stakeholders to anticipate market fluctuations, manage risks, and optimize investment strategies \cite{staum2002simulation,chan2015simulation,lopez2016review}. The growing trend of applying machine learning on price feature data in financial research \cite{rundo2019machine} leads to the need for a simulator that can generate the data with high-fidelity market features. On the other hand, the existing research indicates that there is a non-stationary context in the market \cite{michael2003financial,hens2009handbook}. Leveraging the semantic context as control, the market simulation could be more explainable, controllable, and diverse, benefiting downstream applications \cite{purkayastha2012diversification}. This further leads to the need for a context-aligned, high-fidelity market feature generator. However, existing simulators focus on generating Limit Order Book with agent-based model \cite{samanidou2007agent,axtell2022agent} and generative model \cite{takahashi2019modeling}. While ensemble methods, for example, a multi-expert system can be combined with time-series generative models \cite{mogren2016c,yoon2019time,ni2020conditional} to generate aligned market feature data, this solution could be complex and inefficient \cite{9893798}.  

To address the problem, we proposed Market-GAN, a controllable generator with semantic context for financial simulation of the market features. Our key contributions are: i) we construct the Contextual Market Dataset with the stock ticker, history state, and market dynamics extracted by a market dynamics modeling algorithm as semantic context, addressing the absence of a financial dataset with context; ii) we propose Market-GAN, an innovative hybrid architecture which is the first contextual generative model for financial market features; iii) we design a two-stage training scheme of pre-training and adversarial training for a better initialization of the generator, to address the mode collapse \cite{wiatrak2019stabilizing} observed in training complex GAN networks; iv) with a discussion of the evaluation for financial market simulation, we conduct comprehensive experiments on the generated data with the metrics from the perspective of context alignment, fidelity, data usability, and market facts. Market-GAN showcases superior performance compared with 4 benchmark methods.

\section{Background and Related Works}

\noindent\textbf{Financial Market Simulator.}  Over the years, simulators have emerged as a valuable tool for studying the behavior of financial markets in a controlled environment. Agent-based model methods are widely applied to simulate the Limit Order Book of financial markets \cite{samanidou2007agent,axtell2022agent}. Recent progress has combined the agent-based model with stochastic models \cite{shi2023neural}. While agent-based models offer insights by simulating individual agent behaviors, they rely heavily on behavior models of agents and empirical market models, which sheds some doubts on the plausibility of using this method to simulate complex market \cite{gould2013limit, PhysRevE.76.016108,vyetrenko2020get}. While the price feature is an important data source, especially in fundamental and technical analysis as illustrated by \cite{petrusheva2016comparative,dechow2001short,gite2021explainable,miao2014high}, its simulation is also essential.

\noindent\textbf{Time-Series Data Generation.}
Generated time-series data can be useful in data augmentation or when real data is scarce or sensitive, especially in financial applications, showing its potential for market simulation. Among all the methods, solutions based on GAN gain popularity in recent years. RCGAN \cite{esteban2017real} introduces RNN with conditional inputs to multi-variant time-series generation via GAN architecture. TimeGAN \cite{yoon2019time} utilizes a two-stage autoencoder and GAN training scheme to learn goal and local goals together. SigCWGAN \cite{ni2020conditional} combines continuous-time stochastic models with its signature metric. Stock-GAN \cite{takahashi2019modeling} is a generative model that generates the order stream instead of the market features. 
While FIN-GAN \cite{takahashi2019modeling} introduces GAN to generate price features, the use of vanilla GAN is rudimentary compared to the benchmark methods of time-series generative models.

\noindent\textbf{Contextual Generation.}  Contextual generation refers to the generation of content that is highly relevant given a certain context. Unlike generic content generation, contextual generation ensures that the produced content aligns with the given semantics of the context. This capability is vital for tasks where precision and relevance are paramount, for example, financial simulation. Conditional GAN \cite{mirza2014conditional} introduces a method to direct the generation process with conditions in GAN architecture. CGMMN \cite{ren2016conditional} enables contextual generation based on GMMN \cite{li2015generative}. More recent works show a greater ability to use semantic context as conditions in various domains, including natural language processing \cite{platanios2018contextual,openai2023gpt4}, and image generation \cite{karras2020analyzing,zhang2023adding}. While there is a blooming trend in the multi-modality generation using text as context \cite{saharia2022photorealistic,ramesh2021zero, crowson2022vqgan, rombach2022high}, this paradigm has not been introduced to financial data generation.

\section{Contextual Market Dataset}

To build the Contextual Market Dataset that addresses the lack of financial datasets with semantic context, we first define the plausible context for the financial market. In the spirit of combining fundamental and technical analysis methods in financial research, we propose a hybrid asset price model $\mathbf{X}_{t,\epsilon}=f(C_{\text{lt}}(t), C_{\text{mt}}(t), C_{\text{st}}(t,\epsilon))$. This model takes into account different time scales and viewpoints to estimate asset prices, which are represented as $\mathbf{X}$ based on three types of context:
i) long-term fundamentals $C_{\text{lt}}(t)$ with a time range of multiple years; ii) mid-term market dynamics $C_{\text{mt}}(t)$ with a range of $2$ to $6$ months; and iii) short-term history with volatility $C_{\text{st}}(t,\epsilon)$ with a range less than $2$ months.

Consider a financial market with a set of financial instruments $\mathcal{L}$ and a historical time range of $T$. We define a consecutive batch of price features in a stock market as $\mathbf{X}\in \mathbb{R}^{T \times F}$, where $T$ is the length of the batch $\mathbf{X}$ and $F$ is the number of features. For a stock market, $\mathbf{X}$ carries the following attributes: i) market dynamics $C_{\text{mt}}(t)=d \in \mathcal{D}$, with $\mathcal{D}$ being the market dynamics space ii) stock ticker $C_{\text{lt}}(t)=l \in \mathcal{L}$, where $\mathcal{L}$ is the stock ticker space iii) near history $C_{\text{st}}(t,\epsilon)=\mathbf{H}_{t,\epsilon}\in \mathbb{R}^{T_{H}\times F_{H}}$, with $T_{H}$ being the length of the history batch and $F_{H}$ being the number of features, and $\epsilon$ is a noise vector sampled from a normal distribution that models the latent volatility in the market. In the Contextual Market Datatset, the real market data $R$ is: 
\noindent\begin{multline}
R=\{(\mathbf{X}_{t,\epsilon_{X_{t}}},\mathbf{H}_{t,\epsilon_{H_{t}}},d_{t},l_{t})| \\ \mathbf{H}\in \mbox{history of }\mathbf{X},d_{t}=d_{X_{t}},l_{t}=l_{X_{t}}\},
\end{multline}
where $\mathbf{X}$ is a batch from the historical data stream maintaining the same dynamics $d$ and stock ticker $l$.

\label{sec:MDM}
\begin{figure}[h!]
\centering
\includegraphics[width=0.99\columnwidth]{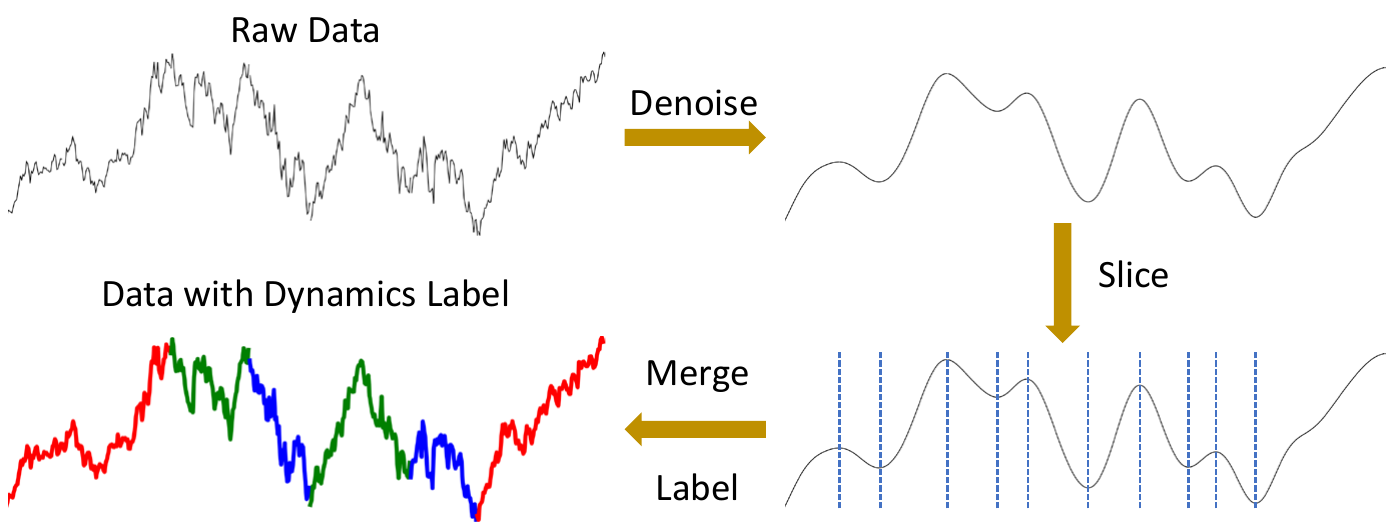} 
\caption{Overview of the market dynamics modeling}
\label{fig:MDM}
\end{figure}
\subsection{Market Dynamics Modeling}
While $\mathbf{H}$ and $l$ have ground truth values associated with each $\mathbf{X}$, the market dynamics $d$ of $\mathbf{X}$ still require mining. 

\begin{algorithm}[b]
\caption{Market Dynamic Modeling}
\label{alg:MDM}
\textbf{Input}: Time-series $x$\\
\textbf{Parameter}: Number of Dynamics $|\mathcal{D}|$, Minimum Length $l_{min}$, Threshold $\theta$, Distance Constraint $c$\\
\textbf{Output}: Dynamics Label $d$
\begin{algorithmic}[1] 
\STATE Denoise $x$ with a low-pass filter into $x^{'}$.
\STATE Slice $x^{'}$ into batches set $B$ by extremums.
\STATE For each batch $b_{i}$ in $B$, merge $b_{i}$ with $b_{i+i}$ $\text{length of } b_{i} \indent <l_{min}$
\WHILE{Clustering not converge}
\STATE Calculate the slope of each batch $b_{i}$ with linear regression models and label their dynamics $d$ into $D$ categories by the percentile of their slope. Merge $b_{i}$ with $b_{i+1}$ if distance$(b_{i},b_{i+1}) < \theta$ and label distance $|d_{i}-d_{i+1}|\le c$
\ENDWHILE

\end{algorithmic}
\end{algorithm}

In Algorithm \ref{alg:MDM}, the Market Dynamics Modeling (MDM) method integrates the clarity of linear regression with the flexibility of clustering which effectively measures the similarity between two time sequences.

As described in Algorithm \ref{alg:MDM}, the data is first denoised by a low-pass filter, a signal processing technique used for the estimation of a desired signal from an observed noisy signal, to filter out the high-frequency volatility. After the data is denoised, we slice it by extremums into batches $B$ of the shortest market trend that is not dividable. Then, they are merged to reach $l_{min}$, which is the minimum expected length of a short-term market structure. With the short-term structures, we cluster them into mid-term dynamics $d$ by using the normalized Euclidean distance to measure the similarity of adjacent batches to find the short-term structures while using the slope calculated from the linear regression model to decide the dynamic label of each $b_{i}$. An overview of the process is illustrated in Fig \ref{fig:MDM}. Detailed hyper-parameters are in Appendix A.

\subsection{Dynamics Modeling Result Analysis}
\setlength{\tabcolsep}{2.5pt}
\begin{table}[h!]
\centering
\small
\begin{tabular}{c|c|c|c|c}
\hline
Dynamic  & $length$ & $s$ & $ul$ & $dl$ \\
\hline
0(bear) & $111(\pm 55)$  & $-2.0(\pm 1.6)$& $23(\pm 17)$ & $80(\pm 42)$   \\
1(flat) & $145(\pm 132)$  & $1.7(\pm 1.0)$ & $108(\pm 125)$ & $25(\pm 12)$  \\
2(bull) & $175(\pm 136)$ & $4.8(\pm 1.8)$  & $156(\pm 134)$ & $19(\pm 17)$ \\
\hline
\end{tabular}
\caption{Modeling result of AAPL. Unit of slopes are $e^{-3}$}
\label{table:Labeling Result of AAPL}
\end{table}
The modeling result of AAPL from 2000 to 2023 is shown in Table \ref{table:Labeling Result of AAPL} with the number of dynamics as 3, minimum length of 50, threshold of 0.03, and distance constraint of 1. We evaluate the average length $length$, average slope $s$ by the linear regression model, average maximum uptrend length $ul$, and maximum downtrend length $dl$ of the three dynamics (bear, flat, bull) we labeled. Results show that i) $l$ falls in the range of a mid-term context of 2 to 5 months; ii) $s$ is aligned with the bear, flat, bull semantic; iii) $ul$ of dynamics 0 and $dl$ of dynamics 2 is far below $l=50$, indicating that the dynamics should not be further segmented, thus they are correctly labeled.

\begin{figure}[h!]
    \centering
    \begin{subfigure}{0.45\columnwidth}
        \centering
        \parbox[c][3cm][c]{\linewidth}{ 
            \includegraphics[width=\linewidth]{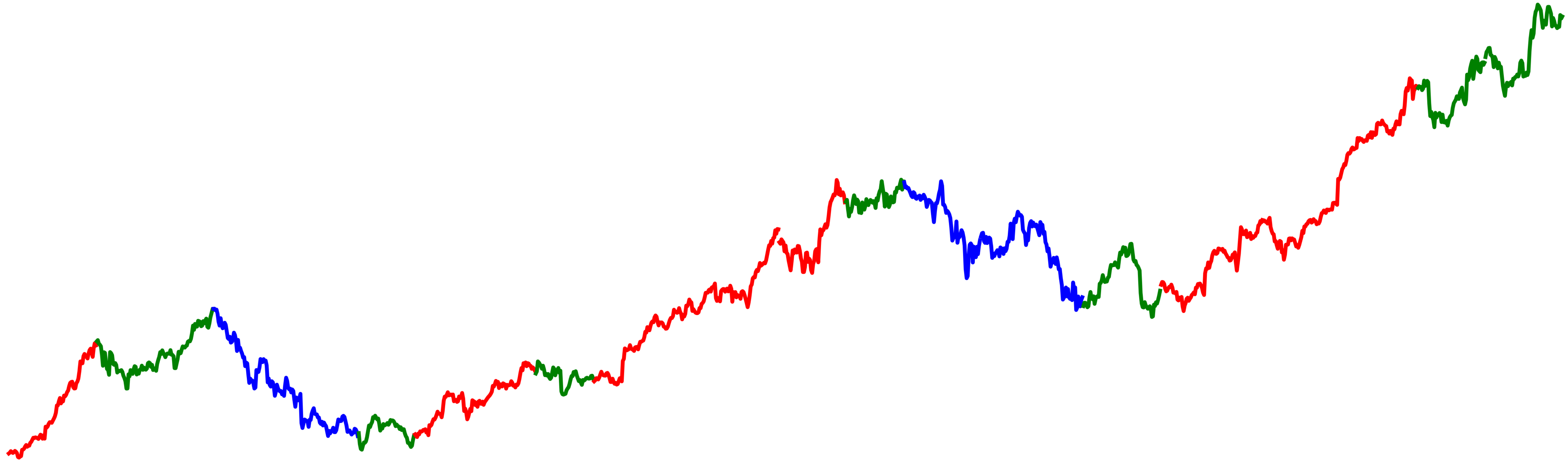}
        }
        \caption{Data with Dynamics}
        \label{fig:MDM_adjclose}
    \end{subfigure}%
    \hfill
        \begin{subfigure}{0.45\columnwidth}
        \centering
        \parbox[c][3cm][c]{\linewidth}{ 
            \includegraphics[width=\linewidth]{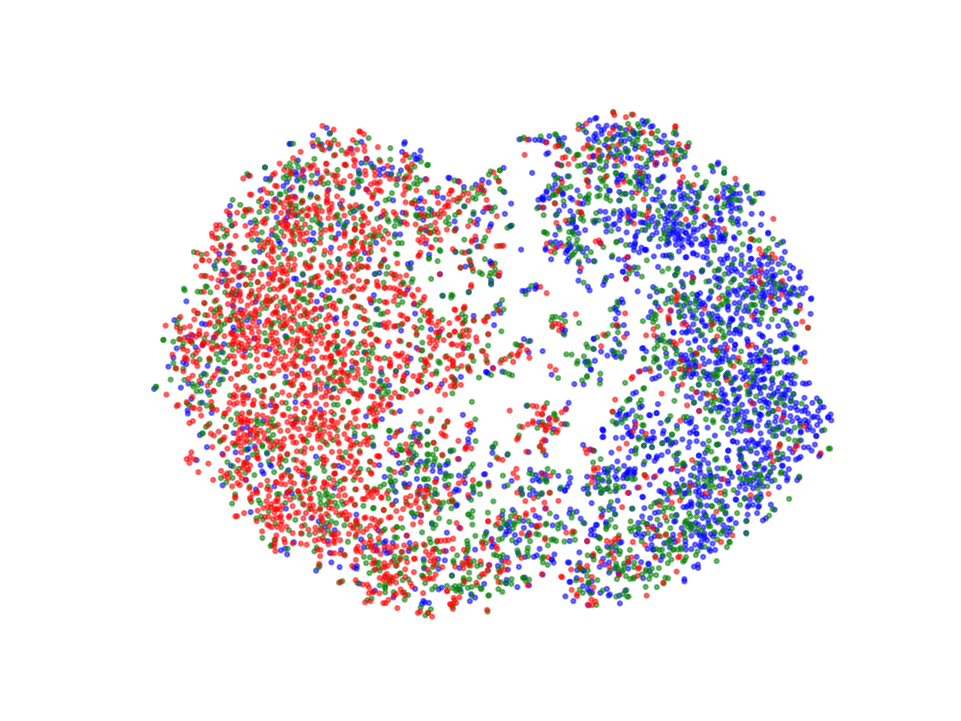}
        }
        \caption{t-SNE plot}
        \label{fig:MDM_TSNE}
    \end{subfigure}%
    \caption{Visulizations result of dynamics modeling where dynamics 0,1,2, is marked as blue, green, and red. (a) illustrates a segment of the Adj close feature of AAPL (1473 days); (b) illustrates the t-SNE plot of 29 stocks (except DOW) in DJI.}
    \label{fig:MDM_result}
\end{figure}
Fig \ref{fig:MDM_result}(a) shows an example of data with the dynamics labeled. The t-SNE plot in Fig \ref{fig:MDM_result}(b) shows data from dynamics 0 and 2 are in distinct clusters, while data from dynamic 1 spreads in between, representing the bear, flat, and bull markets.

\section{Market-GAN}
\label{sec:Market-GAN}
With the Contextual Market Dataset, we describe the generation goal of Market-GAN and its hybrid architecture.

\subsection{Problem Formualtion}
\label{sec:Generation Goals}
As semantic context $\mathbf{H}$, $d$, $l$, and $\mathbf{X}$ form a one-to-one data-context pair, the limited real data $\mathbf{X}$ would lead to a limited context in $R$. Given the non-stationary nature of the financial market, the context distribution $p(\mathbf{H}_{t,\epsilon},d_{t},l_{t})$ may shift over time, leading to Out Of Distribution (OOD) problem with the context of $R$ only encompasses a subset of all the possible context sets.
By utilizing the generative model $G$, we aim to augment $R$ with the generated data
\noindent\begin{equation}
\footnotesize
F=\{( \hat{\mathbf{X}}_{\epsilon}, \mathbf{H}, d, l) | \mathbf{H} \in 	\mathbb{R}^{T_{H} \times F_{H}}, d\in \mathcal{D}, l\in \mathcal{L}, \epsilon \in \mathcal{N}\},
\end{equation}
where $\hat{\mathbf{X}}_{\epsilon}=G(\textbf{Z},\mathbf{H},d,l)$, and $F$ is generalized to any plausible context $\mathbf{H},d,l$. For simplicity, we omit $t$ and $\epsilon$.

We aim to learn a generative model $G(\textbf{Z},\mathbf{H},d,l)=\mathbf{X}_{t,\epsilon}$ that uses semantic context to control the simulation of the market. 
$G$ should learn the distribution $\hat{p}(\mathbf{X}, \mathbf{H}, d, l)$ that approximates $p(\mathbf{X}, \mathbf{H}, d, l)= p(\mathbf{H},d,l) \cdot p(\mathbf{X}|\mathbf{H},d,l)$ so that the generated $F$ can be controlled by semantic context $\mathbf{C}=(\mathbf{H},d,l)$. Given the context $\mathbf{C}$ in the generation, the model needs to learn $\hat{p}(\mathbf{X}|\mathbf{H},d,l)$. With the assumption that $\mathbf{H}$, $d$, $l$ are disentangled in the representation space and thus independently conditioned on $\mathbf{X}$, the objective can be represented as $p(\mathbf{X}|\mathbf{H},d,l)\propto  p(\mathbf{H},d,l|\mathbf{X})=p(\mathbf{H}|\mathbf{X})\cdot p(d|\mathbf{X}) \cdot p(l|\mathbf{X})$.

As most downstream machine learning tasks assume that financial time-series features follow the Markov property, our model also learns the auto-regressive transaction distribution $p(\mathbf{X}_{t}|\mathbf{X}_{0:t-1})$, where $\mathbf{X}_{t}$ is the t-th tick of $\mathbf{X}$ and $\mathbf{X}_{0:t-1}$ is its preceding feature batch. Specifically, since $d$ and $l$ have different semantics compared with $\mathbf{X}$ and $\mathbf{H}$, the auto-regressive transaction distribution can be expressed as $p(\mathbf{X}_{t}|\mathbf{X}_{0:t-1},\mathbf{H},d,l)$.

\begin{figure*}[h!]
\centering
\includegraphics[width=0.90\textwidth]{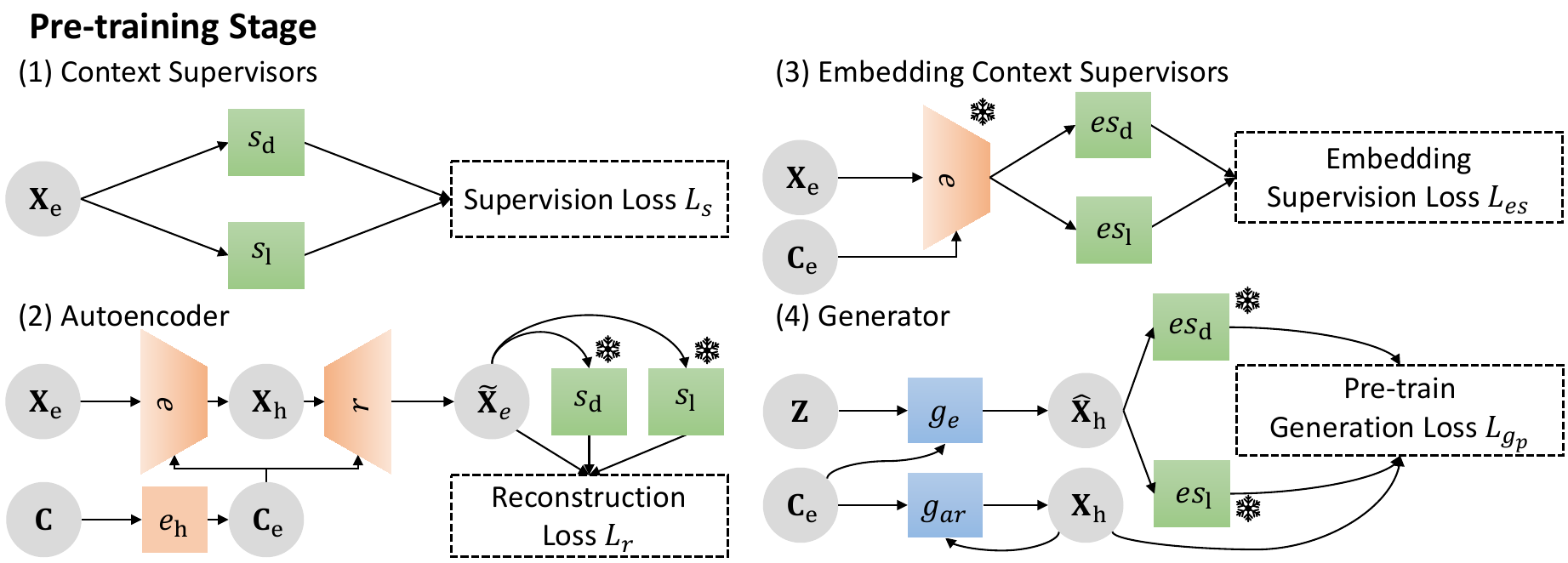} 
\caption{Training scheme of Market-GAN in the pre-training stage. The snowflake indicates the network parameters are frozen in the stage.}
\label{fig:Pre-train}
\end{figure*}

\subsection{Architecture}

We follow the autoencoder and GAN structure of TimeGAN \cite{yoon2019time} and enhance it with data transformation,  supervisor teachers, and C-TimesBlock. The raw data $\mathbf{X}$ and context $\mathbf{C}$ is transformed into the encoding space $\mathbf{X}_{e}$ and $\mathbf{C}_{e}$ by the transformation layer and then embedded into the latent space $\mathbf{X}_{h}$ with an embedding network $e$. We train context classifiers $s_{d}$, $s_{l}$ in encoding space and $es_{d}$, $es_{l}$ in the latent space to capture $p(d|\mathbf{X}_{e})$, $p(l|\mathbf{X}_{e})$, $p(d|\mathbf{X}_{h})$, $p(d|\mathbf{X}_{h})$ and transfer this knowledge to the generator $g$. The generator $g$ is a combination of embedding generator $g_{e}$ and auto-regression generator $g_{ar}$ where $g(\mathbf{Z},\mathbf{C})=g_{ar}(g_{e}(\mathbf{Z},\mathbf{C}_{e}),\mathbf{C}_{e})$. With the discriminator $dis$, the generator learns to generate the latent $\hat{\mathbf{X}}_{h}$, while the decoder $r$ transforms it to $\hat{\mathbf{X}}_{e}$ and the inverse transformation layer transforms the encoding back to the market features $\hat{\mathbf{X}}$.

\subsection{Data Transformation}

The Data Transformation layer transforms the input $\{(\mathbf{X},\mathbf{H},d,l)\}$ to an encoded feature $\{(\mathbf{X}_{e},h_{e},d_{e},l_{e})\}$. The price features typically follow these patterns: i) key information is carried by deviation, which is relatively small compared to the feature value; ii) the distribution shift accumulates over time, and iii) price features have correlations, causing the following issues in learning: i) the immersion of deviation in the raw data; ii) ineffective data normalization on non-Gaussian long periods of financial data; iii) With $Open$, $High$, $Low$, $Close$ (OHLC) as price features, violation of the $Low \leq (Open, Close) \leq High$ constraint would make data harmful to downstream tasks. We propose a data transformation layer to resolve these issues. The feature encoding layer reparameters the OHLC features into $Low$,$Open-Low$,$Close-Low$,$High-\text{max}(Open, Close)$ such that all features of $\mathbf{X}_{e}$ and $\mathbf{H}_{e}$ have non-negative deviations. Then, the $Low$ feature is processed into step-wise differential $Low_{d}$. The near history $\mathbf{H}_{e}$ is then used to normalize itself and $\mathbf{X}_{e}$ to mitigate the distribution shift and simultaneously prevent the leakage of future information to each sample. The embedding network $e_{h}$ further compress $\mathbf{H}_{e}$ into $h_{e}$. For simplicity, we let $\mathbf{C}_{e}=\{h_{e},d_{e},l_{e}\}$.

\subsection{C-TimesBlock}
We develop a C-TimesBlock (CTB) that balances time series representation learning and condition modulation. While RNN performs well in a wide range of GAN applications, solely using the RNN in our contextual generation task will result in mode collapse \cite{wiatrak2019stabilizing}, a known challenge in training GANs where the generator produces a limited variety of samples. 
The TimesBlock \cite{wu2023timesnet} casts the input into a 2D spectrum and learns with the Inception Blocks, capturing multi-scale patterns of different time periods. By incorporating RNN with TimesBlock, the C-TimesBlock captures the temporal dependency in both 1D and 2D space, mitigating mode collapse. The context alignment scores in our experiment show the superior performance of C-TimesBlock over RNN on context alignment.

\section{Training Scheme}
\label{sec:trainingscheme}
Training the generator $g$ from scratch for both context alignment and fidelity generation is challenging, leading to a mode collapse. To alleviate this issue, we design a two-stage training scheme for Market-GAN.

\subsection{Pre-training Stage}
Using an autoencoder to extract representation and context supervisors as teachers for alignment, the pre-training stage prepares the generator with a better initialization in adversarial training with discriminator $dis$ as shown in Fig \ref{fig:Pre-train}. \\
\noindent\textbf{Context Supervisors.}
\label{sec:Condition Supervisor Pre-training}
We adopt TimesNet \cite{wu2023timesnet} for context supervisors $s_{d}$ and $s_{l}$. The objective is to reduce the supervision loss $L_{s}(\mathbf{X}_{e})$:
\noindent\begin{equation}
\underset{\theta_{s_{d}},\theta_{s_{l}}}{\text{min}}CE(d_{e},s_{d}(\textbf{X}_{e}))+CE(l_{e},s_{l}(\textbf{X}_{e})),
\end{equation} where $CE$ is the cross-entropy loss. 

\noindent\textbf{Autoencoder.} We undertake the training of the embedding network $e$, the embedding decoder $r$, and the history embedding network $e_{h}$ during this phase. While the embedding network learns the latent representation $\mathbf{X}_{h}=e(\mathbf{X}_{e},\mathbf{C}_{e})$, the encoded data is reconstructed by $r$ with  $\Tilde{\mathbf{X}}_{e}=r(\mathbf{X}_{h},\mathbf{C}_{e})$. The objective of this phase is to minimize the reconstruction loss $L_{r}(\mathbf{X}_{e})$:
\noindent\begin{equation} \label{eq:autoencoderPretrain}
\footnotesize
\underset{\theta_{e},\theta_{r},\theta_{e_{h}}}{\text{min}}
||\Tilde{\mathbf{X}}_{e}-\mathbf{X}_{e}||+
\gamma(CE(d_{e},s_{d}(\Tilde{\mathbf{X}}_{e}))+CE(l_{e},s_{l}(\Tilde{\mathbf{X}}_{e}))),
\end{equation}
where $\gamma$ is the context alignment weight. By incorporating classification loss into the training objective of the autoencoder, $e$ derives a latent space that emphasizes a robust contextual representation. While $e$ and $r$ together constitute an autoencoder that transform $\mathbf{X}_{e}$ into the latent $\mathbf{X}_{h}$, considering that  $\mathbf{X}_{e}$ is a 2D return of the $Low$ feature, the autoencoder can be considered as a factor model \cite{duan2022factorvae}.

We train $e_{h}$ from scratch in this phase instead of copying the parameter from $e$ because the outputs of $e_{h}$ and $e$ possess different semantic levels. To curb training instability, we freeze $e_{h}$ in the subsequent training process. 

To achieve data reconstruction within the re-parameterized range, we apply $prelu$ activation to the $Low_{d}$ and $relu$ activation to the remaining three features.

\noindent\textbf{Embedding Context Supervisors.}
With the $\mathbf{X}_{h}$ from pre-trained $e$, we train the embedding dynamics supervisor $es_{d}$, embedding stock ticker supervisor $es_{l}$. The objective is to minimize the embedding supervision loss $L_{es}(\mathbf{X}_{h})$:
\noindent\begin{equation}
\underset{\theta_{es_{d}},\theta_{es_{l}}}{\text{min}}CE(d_{e},es_{d}(\textbf{X}_{h}))+CE(l_{e},es_{l}(\textbf{X}_{h})).
\end{equation}
\noindent\textbf{Generator.} The two components of the generator $g$ are trained with their respective objective. With $\hat{{\mathbf{X}}}_{h}=g_{e}(\textbf{Z},\mathbf{C}_{e})$, the embedding generator $g_{e}$ is trained with the supervision of $es_{d}$ and $es_{l}$ to align with the context. The auto-regression generator $g_{ar}$ is tasked with learning the distribution $p(\mathbf{X}_{h[t]}|\mathbf{X}_{h[0:t-1]}, \mathbf{C}_{e})$ with the real $\mathbf{X}$. The objective is to minimize the pre-train generation loss $L_{g_{p}}$:
\noindent\begin{multline}
    \label{eq:5_GeneratorPretraining}
\underset{\theta_{g_{e}},\theta_{g_{ar}}}{\text{min}}CE(d_{e},es_{d}(\hat{\mathbf{X}}_{h}))+
CE(l_{e},es_{l}(\hat{\mathbf{X}}_{h}))\\+
MSE(\textbf{X}_{h[1:t]},g_{ar}(\textbf{X}_{h[0:t-1]},\mathbf{C}_{e})),
\end{multline} 
where $MSE$ is the mean-squared error loss.

\begin{figure}[h!]
\centering
\includegraphics[width=0.90\columnwidth]{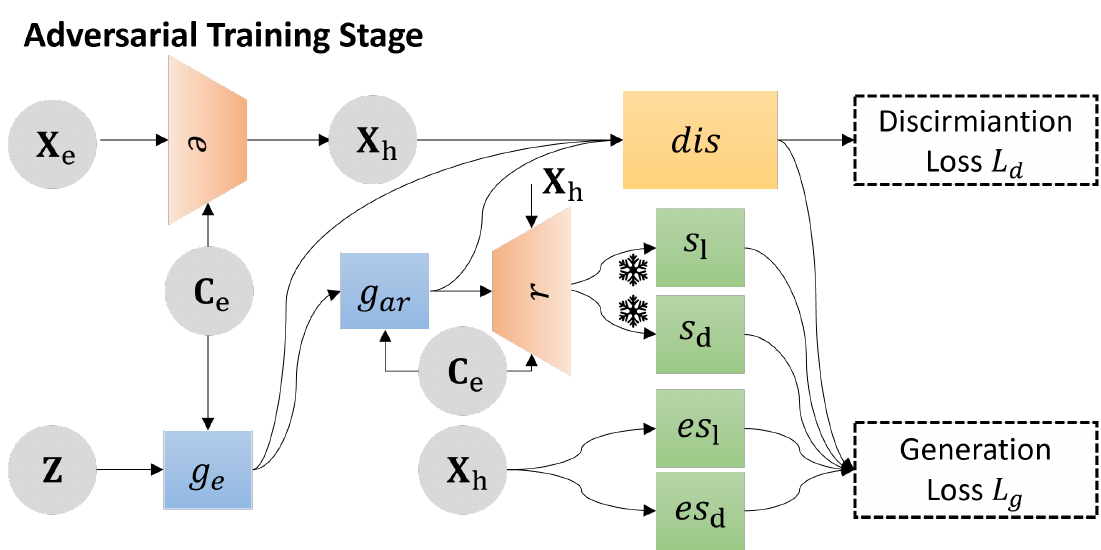} 
\caption{Training scheme of Market-GAN in the adversarial training stage. The snowflake indicates the network parameters of $s_{l}$, $s_{d}$ are frozen in the stage.}
\label{fig:Adversarial_training}
\end{figure}

\subsection{Adversarial Training Stage}

As shown in Fig \ref{fig:Adversarial_training}, we train the generator $g$ and discriminator $dis$ adversarially in this phase with the autoencoder and the supervisors. The discrimination loss is
\noindent\begin{equation}
L_{d}=CE(\mathbf{1},\mathbf{X}_{h})+CE(\mathbf{0},\hat{\mathbf{X}}_{h})+CE(\mathbf{0},\hat{\mathbf{X}}_{h[T+1]}),
\end{equation}
where $\hat{\mathbf{X}}_{h[T+1]}=g_{ar}(\hat{\mathbf{X}}_{h},\mathbf{C}_{e})$. The generation loss is 
\noindent\begin{multline}
\footnotesize
L_{g}=CE(\mathbf{1},d(\hat{\mathbf{X}}_{h}))
+CE(\mathbf{1},d(\hat{\mathbf{X}}_{h[T+1]}))+\\ 
\gamma (L_{s}(\hat{\mathbf{X}}_{e})+L_{es}(\mathbf{X}_{h}))+L_{r}(\mathbf{X}_{e}),
\end{multline}
where we reconstruct $\hat{\mathbf{X}}_{e}$ by $\hat{\mathbf{X}}_{e}=r(\hat{\mathbf{X}}_{h},\mathbf{C}_{e})$, training the Market-GAN with multiple losses jointly. The training objectives for generation and discrimination are
\noindent\begin{equation}
\underset{\theta_{g_{e}},\theta_{g_{ar}},\theta_{r},\theta_{e},\theta_{es_{l}},\theta_{es_{d}}}{\text{min}}\underset{\theta_{d}}{\text{max}}
L_{g}, \quad \underset{\theta_{d}}{\text{min}}L_{d}.
\end{equation}

While adversarial training is challenging with the complex architecture and multi-tasking objectives, our C-TimesBlock and the multi-stage training scheme help to stabilize the process.
With the two-stage training scheme, we train the generative model $G(\textbf{Z},\mathbf{H},d,l)=I(r(g_{ar}(g_{e}(\textbf{Z}, \mathbf{C}), \mathbf{C}), \mathbf{C}))$, where $I$ is the inverse transformation layer.

\section{Evaluating Generated Data with Context}

The main challenge for proposing a robust evaluation metric for financial data is that, unlike images that have an explicit and almost one-to-one explicit, discrete, and fixed semantic of context, the semantics of financial context can be implicit, continuous, and non-stationary.  
We discuss the evaluation metrics for a contextual generative model of financial data with these research questions:
\begin{enumerate}
    \item Is generated $F$ aligned to the given context $\mathbf{C}$ while resembling real data $R$?
    \item Can $F$ enhance the performance of downstream tasks?
    \item Will $F$ be identified as fake data with definiteness?
\end{enumerate}

\noindent\textbf{Experiments Setting.} For the real data $R$, we generate a corresponding synthetic dataset $F$, maintaining the same context $\mathbf{C}$. For each instance of $\mathbf{X}$, we generate a $\dot{\mathbf{X}}$ with a random $\textbf{Z}$, ensuring that $R$ and $F$ have equivalent sizes. While $F$ could be configured to any size, we opt to maintain equality in size between $\mathbf{X}$ and $F$ during evaluation to uphold the fairness of the contrast experiment. We conducted the experiments on a 4090 GPU. Detailed descriptions of the training setups can be found in the Appendix.

\noindent\textbf{Context Alignment.} Following the spirit of using CLIP score \cite{hessel2021clipscore} for evaluation of context alignment, we utilized pre-trained condition supervisors $s_{d}$ and $s_{l}$ to classify the $d$ and $l$ of $F$ with cross-entropy loss $L_{d}=CE_{s_{d}}(F)$ and $L_{l}=CE_{s_{l}}(F)$ as metrics. 

\noindent\textbf{Generation Fidelity.} Following the spirit of using FID \cite{heusel2017gans} to evaluate the fidelity of generated images, the fidelity of $F$ is evaluated using discriminators $d_{e}$, trained via TimeNets. The objective is defined as:
\noindent\begin{equation} \label{eq:posthocD}
\underset{\theta_{d_{e}}}{\text{min}} CE(\mathbf{1},R)+CE(\mathbf{0},F).
\end{equation}
\noindent
The accuracy of $d_{e}$ should be $50\%$ on the test set if it is not distinguishable. Hence, the discrepancy between $R$ and $F$ is evaluated by $L_{D}=|accuracy_{d_{e}} - 50|$. To give $d_{e}$ stronger discrimination ability, we train multi-expert $d_{e(i,j)}$ where each discriminator is trained to classify a subset of data where $d=i$ and $l=j$ and use the average $L_{D}$ of all experts as the metric with 50 training epochs.

\noindent\textbf{Market Facts.}
We evaluate if the generated data adheres to the facts of $OHLC$. $L_{f}$ is the percentage of generated data violating the $Low \leq (Open, Close) \leq High$. Any data that contradicts this fact is definitively identified as fake.

\noindent\textbf{Data Usability.}
With the real set $R$ and the generated set $F$, we have the augmented set $R+F$. In the spirit of TSTR \cite{esteban2017real}, the usability of the data is assessed by examining the performance of the one-step prediction task using the augmented training set, where we applied four prominent time-series forecasting models: TimesNet, TCN, LSTM, and GRU. For a fair comparison, we half the training epoch when using $R+F$ as a training set compared with training only with $R$. We train multi-expert predictors $p_{e(i,j)}$ where each predictor is trained on a subset of data where $d=i$ and $l=j$, and test $p_{e(i,j)}$ on the respective test set of real data. We calculate the Symmetric Mean Absolute Percentage Error (SMAPE) losses of prediction value to evaluate the usability of $F$ in improving the downstream task's performance. The TimesNet predictors are trained with 50 epochs, while TCN, LSTM, and GRU predictors are trained with 200 epochs.

\begin{figure*}[h!]
    \centering
    \begin{subfigure}{0.148\textwidth}
        \centering
        \includegraphics[width=\linewidth]{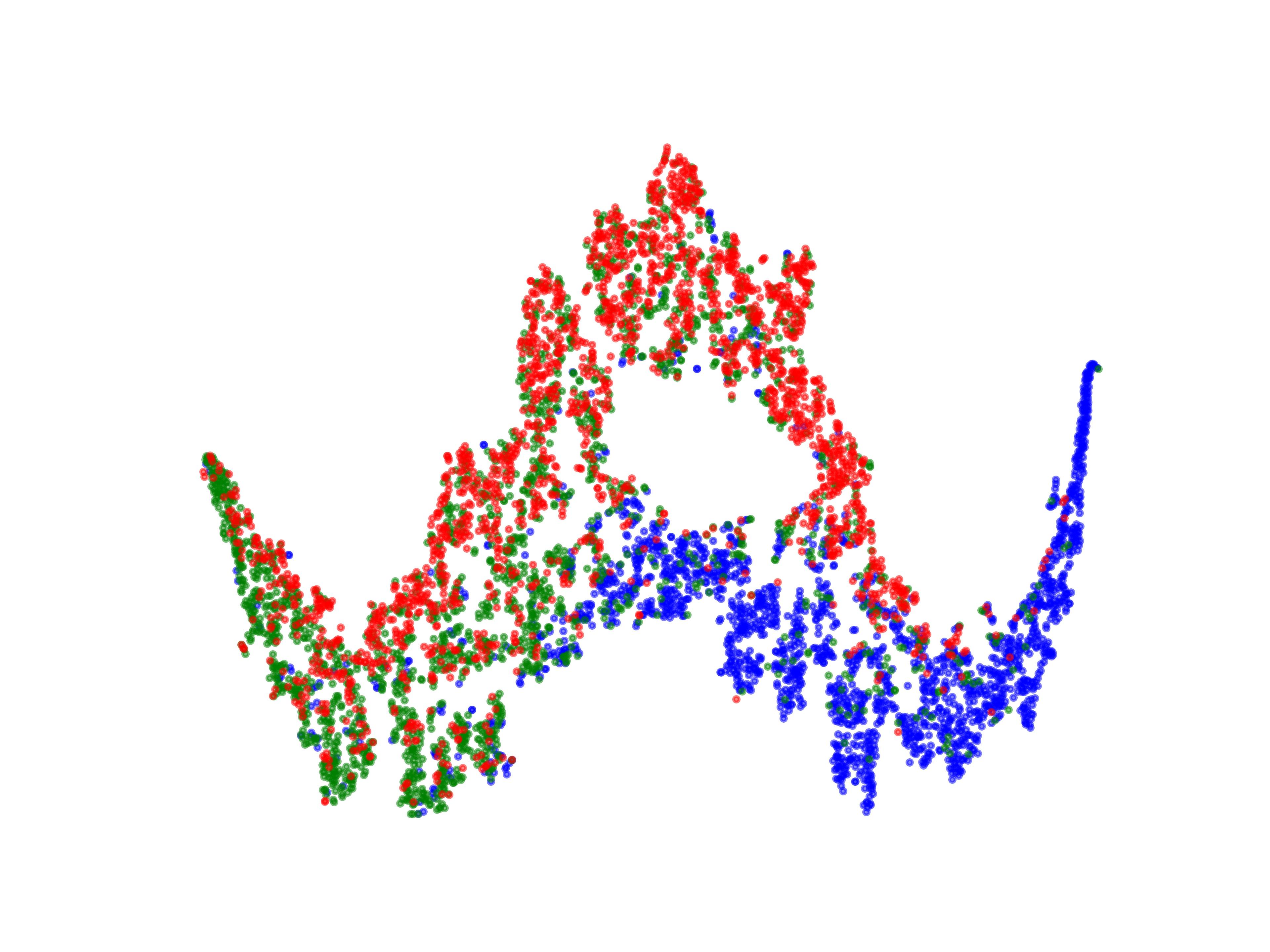}
        \captionsetup{font=small}
        \caption{Market-GAN}
        \label{fig:sub1}
    \end{subfigure}%
    \begin{subfigure}{0.148\textwidth}
        \centering
        \includegraphics[width=\linewidth]{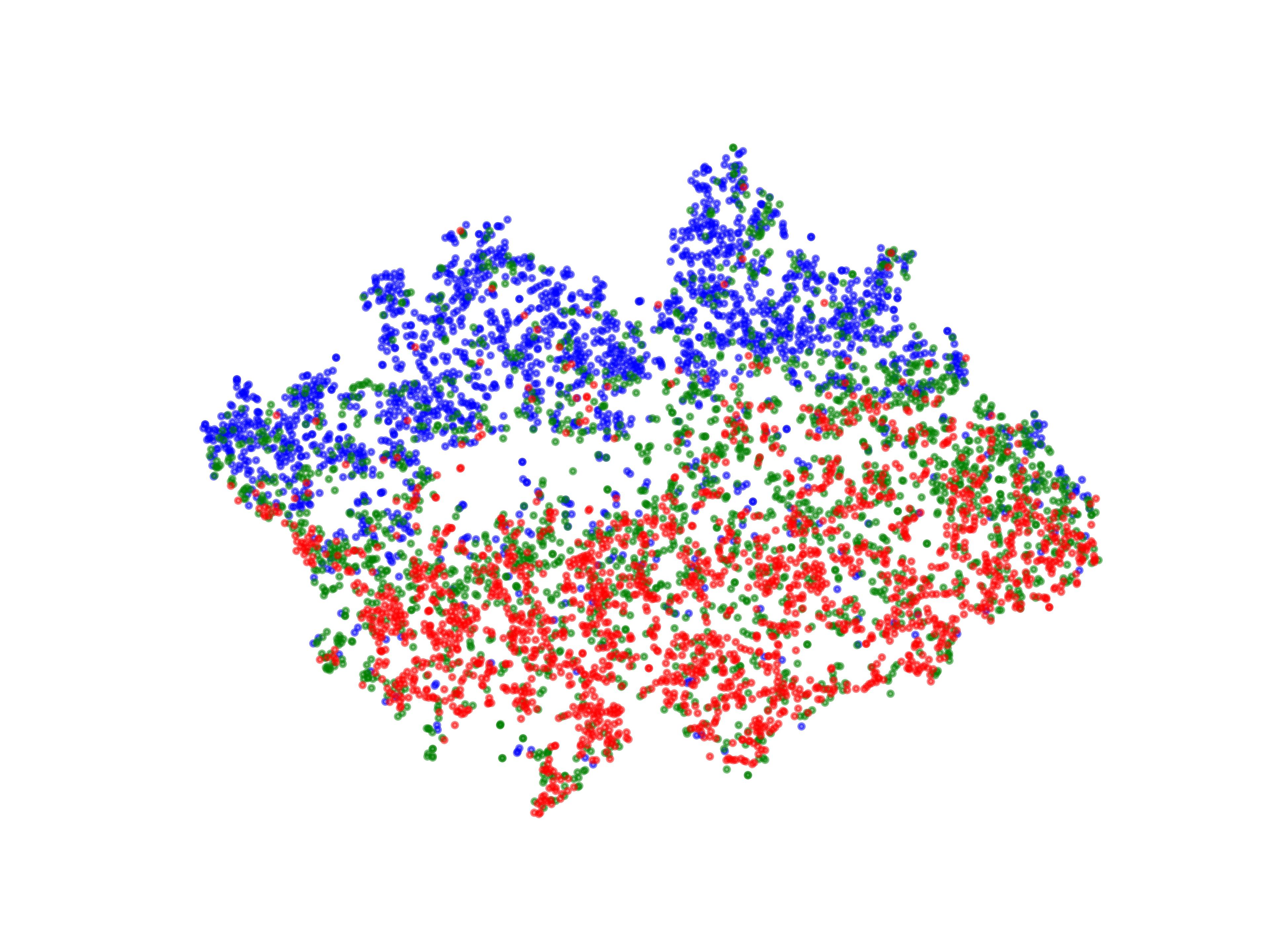}
        \captionsetup{font=small}
        \caption{TimeGAN}
        \label{fig:sub2}
    \end{subfigure}%
    \begin{subfigure}{0.148\textwidth}
        \centering
        \includegraphics[width=\linewidth]{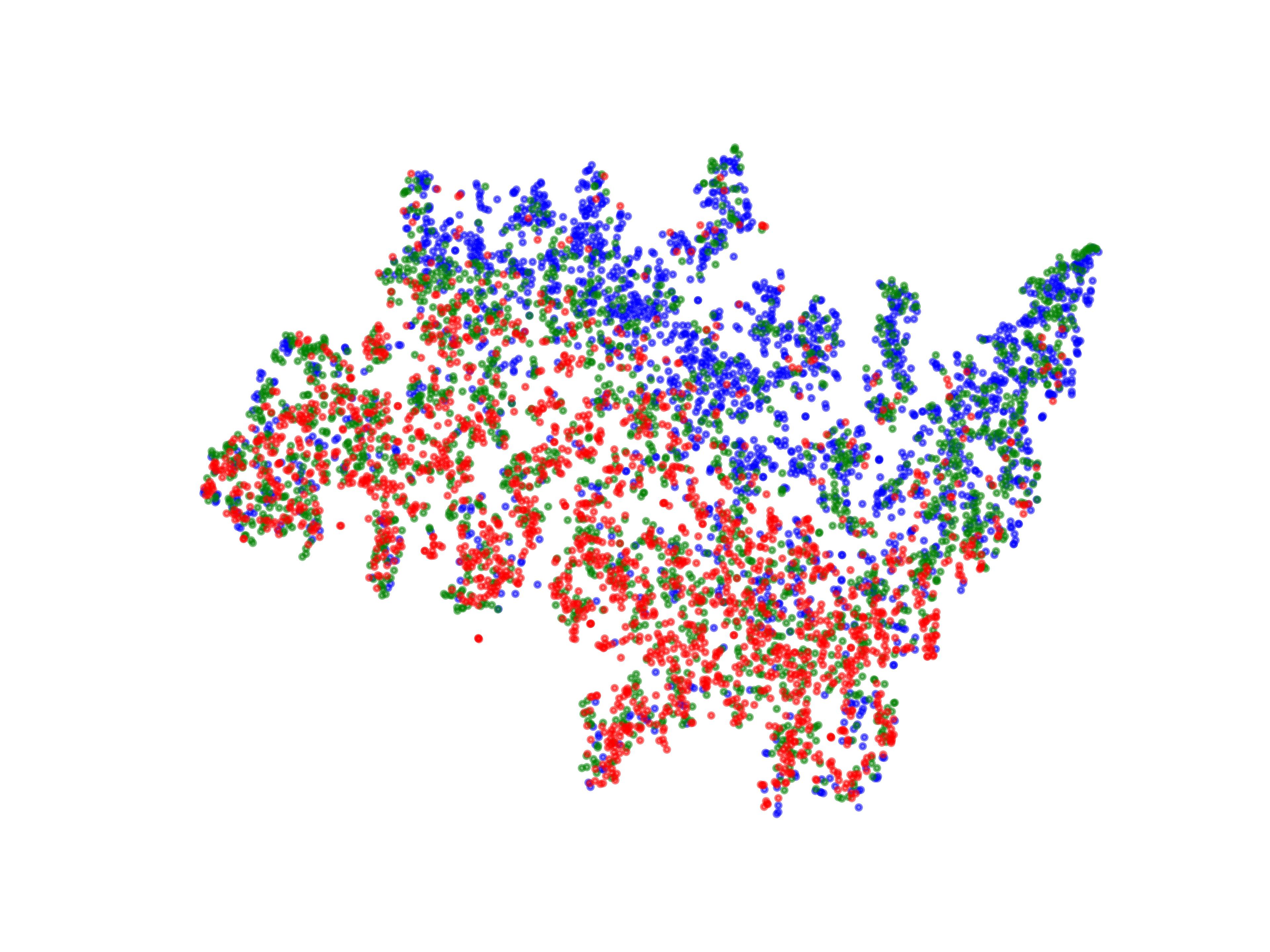}
        \captionsetup{font=small}
        \caption{RCGAN}
        \label{fig:sub5}
    \end{subfigure}%
    \begin{subfigure}{0.148\textwidth}
        \centering
        \includegraphics[width=\linewidth]{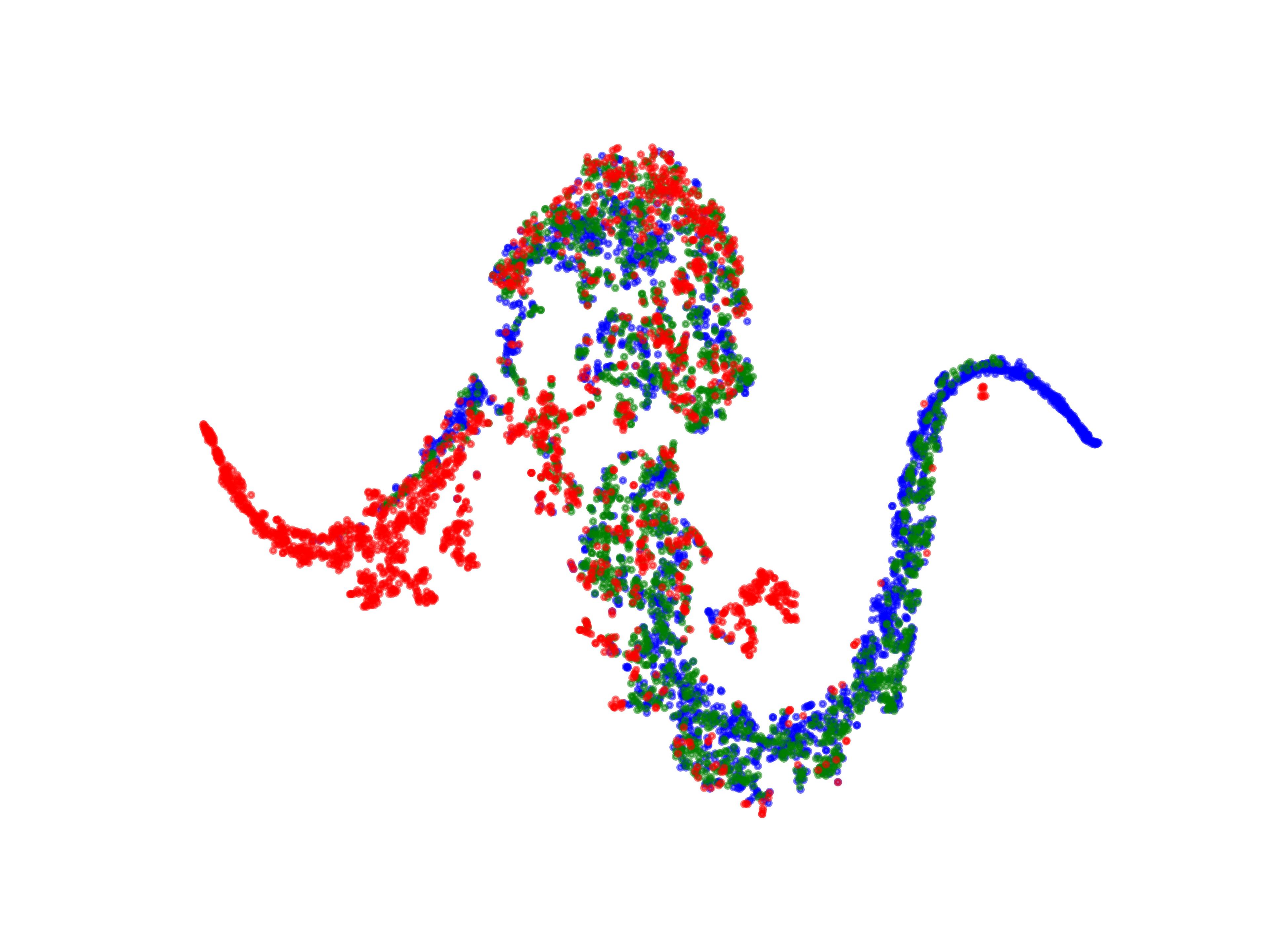}
        \captionsetup{font=small}
        \caption{CGMMN}
        \label{fig:sub3}
    \end{subfigure}%
    \begin{subfigure}{0.148\textwidth}
        \centering
        \includegraphics[width=\linewidth]{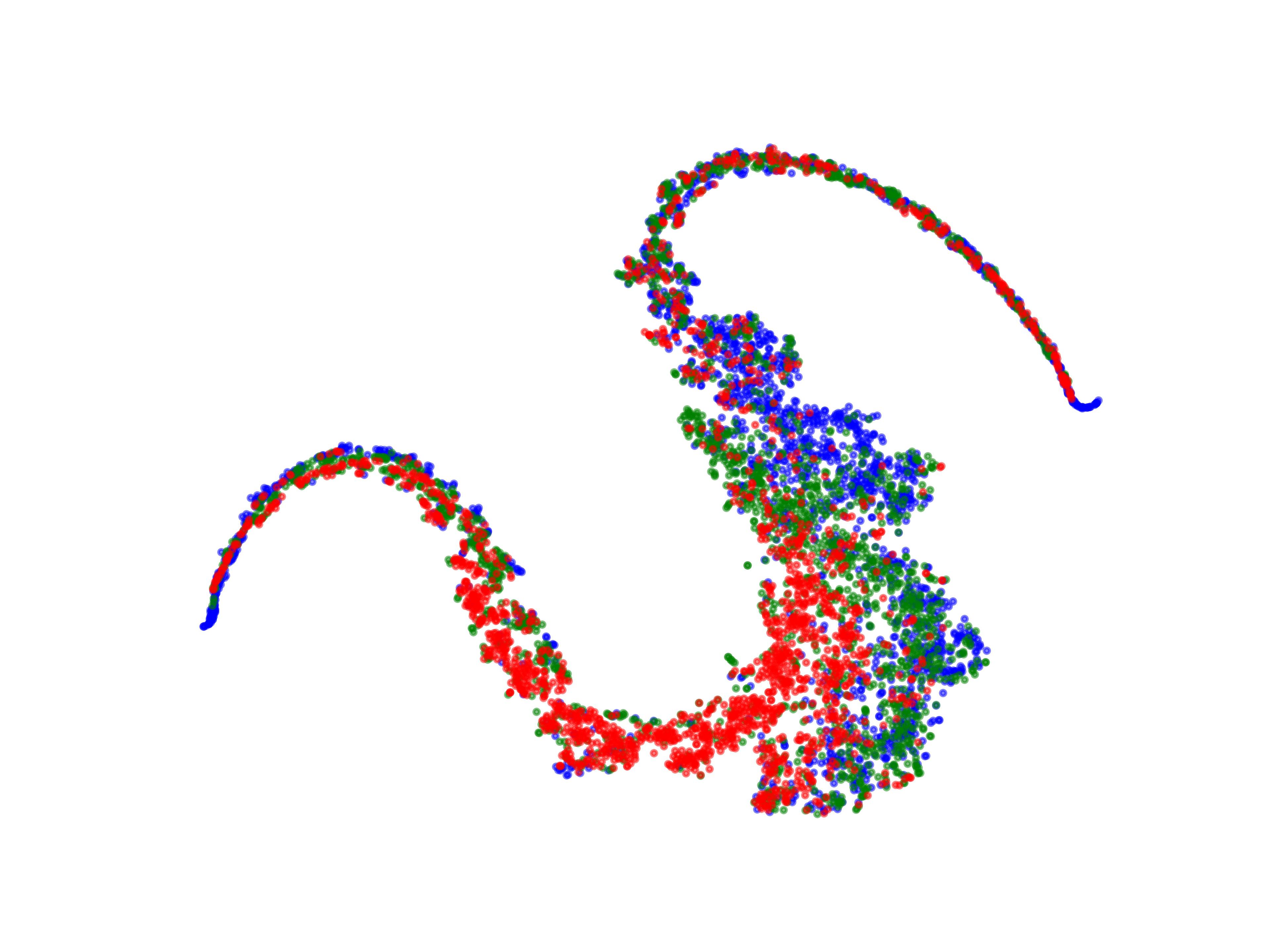}
        \captionsetup{font=small}
        \caption{SigCWGAN}
        \label{fig:sub4}
    \end{subfigure}

    \begin{subfigure}{0.148\textwidth}
        \centering
        \includegraphics[width=\linewidth]{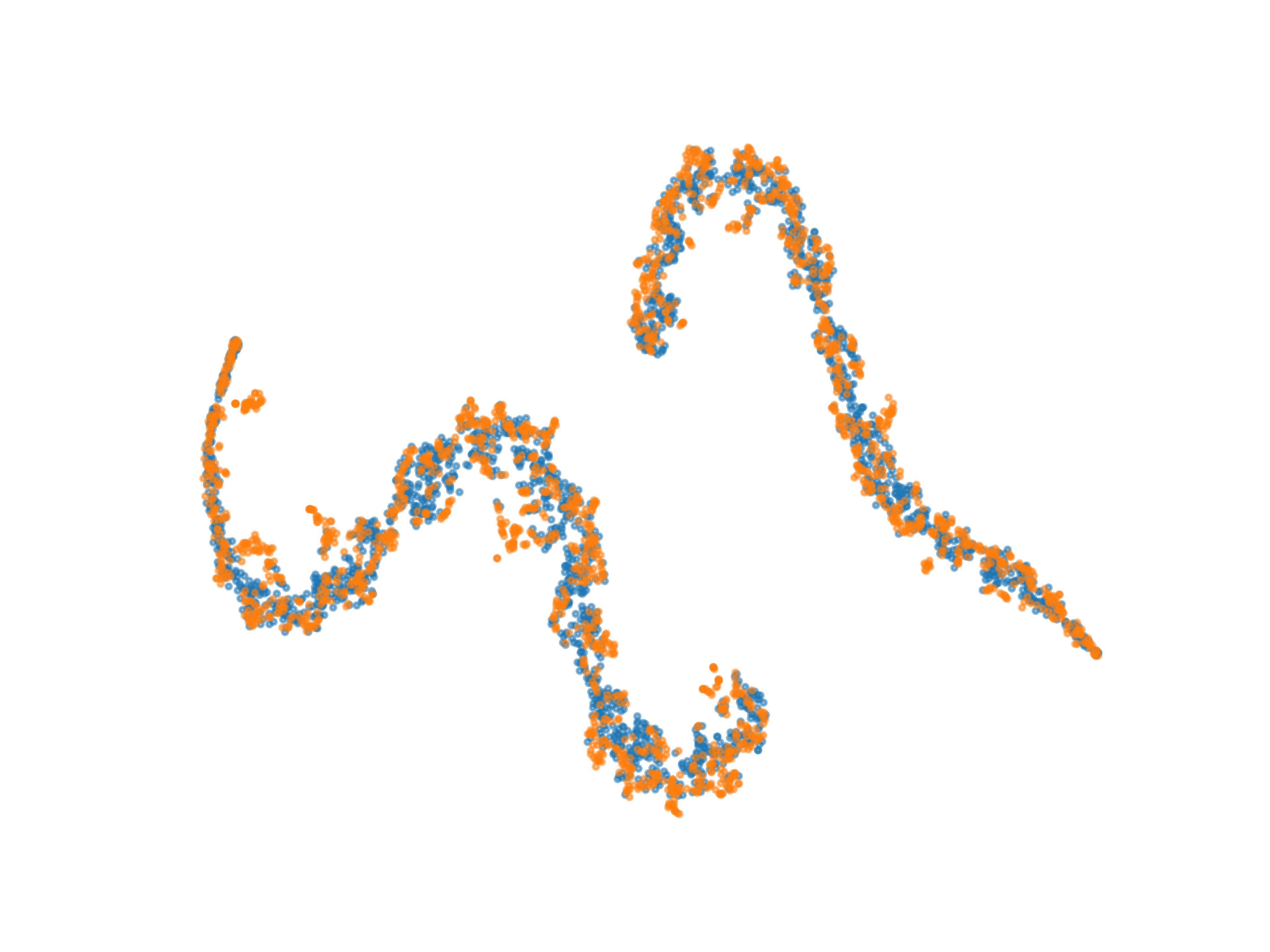}
        \captionsetup{font=small}
        \caption{Market-GAN}
    \end{subfigure}%
    \begin{subfigure}{0.148\textwidth}
        \centering
        \includegraphics[width=\linewidth]{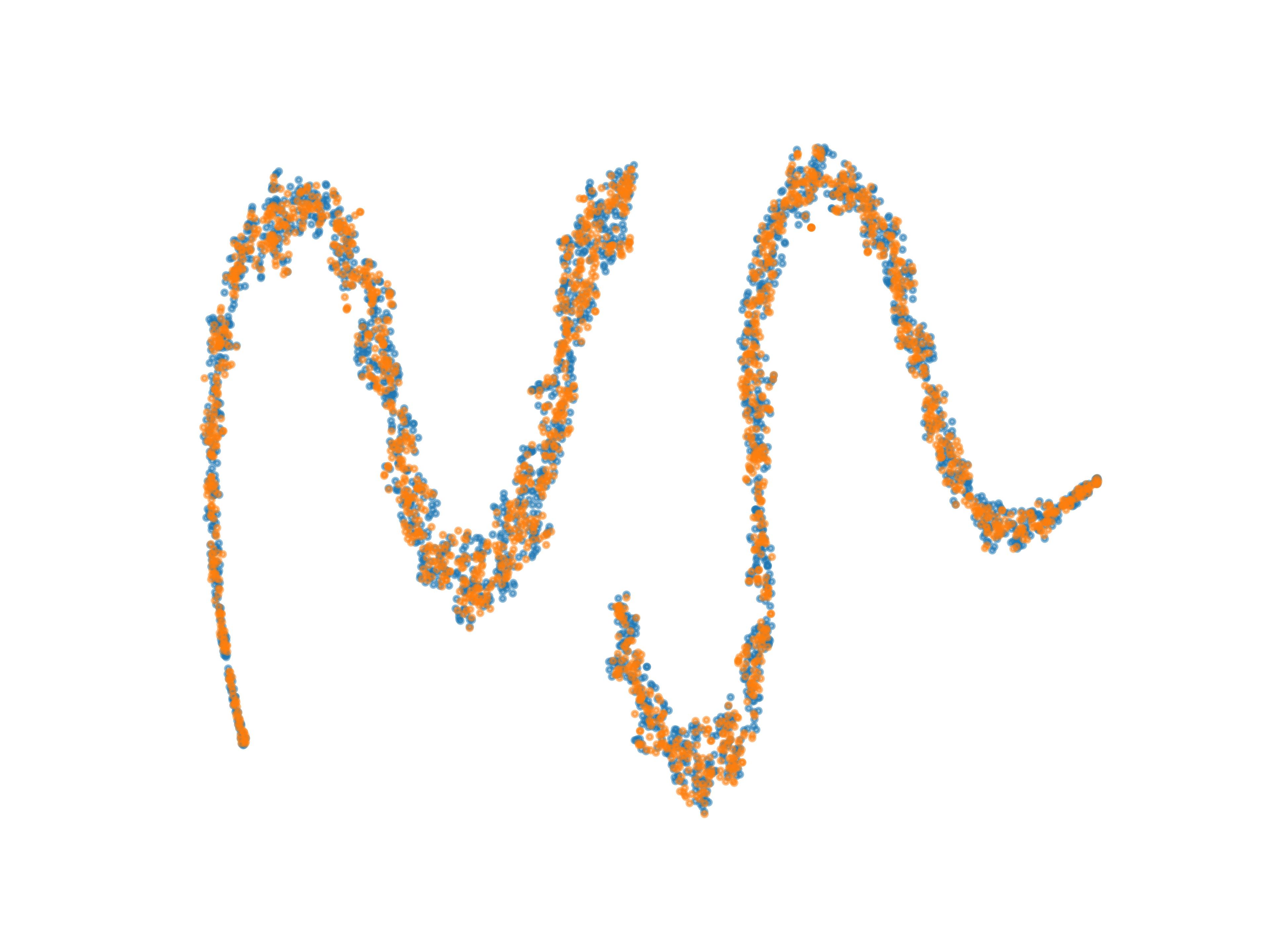}
        \captionsetup{font=small}
        \caption{TimeGAN}
    \end{subfigure}%
    \begin{subfigure}{0.148\textwidth}
        \centering
        \includegraphics[width=\linewidth]{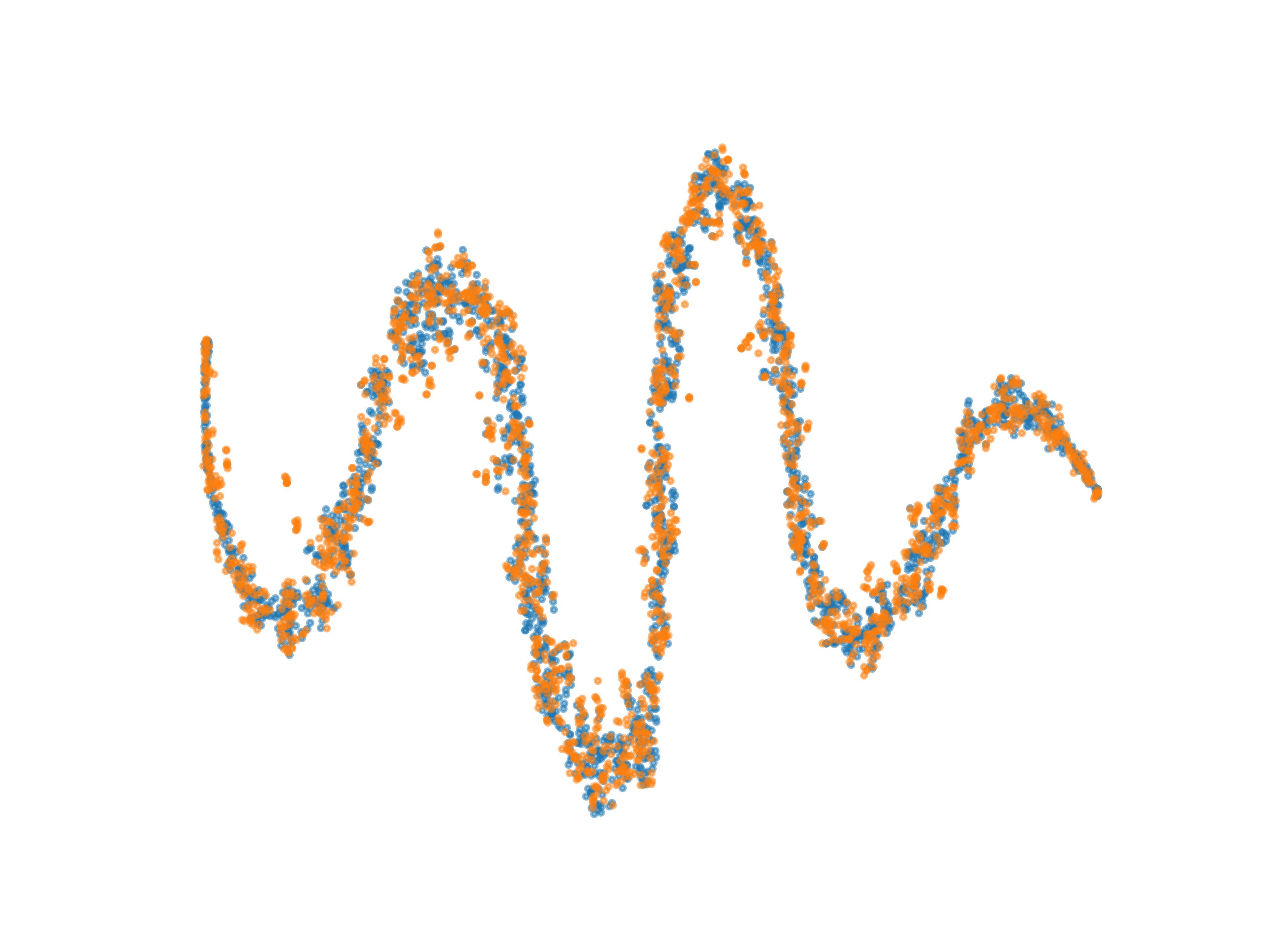}
        \captionsetup{font=small}
        \caption{RCGAN}
    \end{subfigure}%
    \begin{subfigure}{0.148\textwidth}
        \centering
        \includegraphics[width=\linewidth]{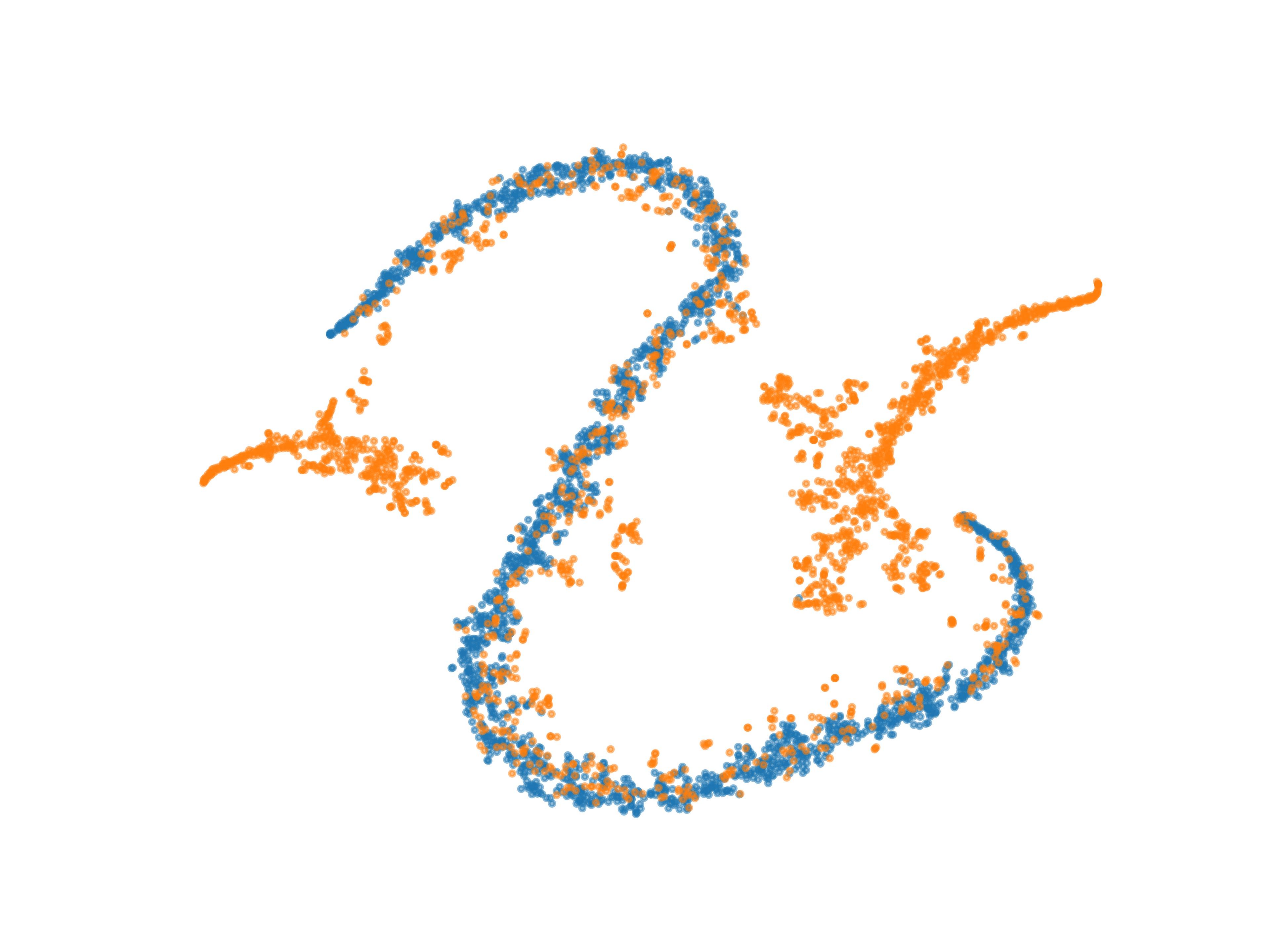}
        \captionsetup{font=small}
        \caption{CGMMN}
    \end{subfigure}%
    \begin{subfigure}{0.148\textwidth}
        \centering
        \includegraphics[width=\linewidth]{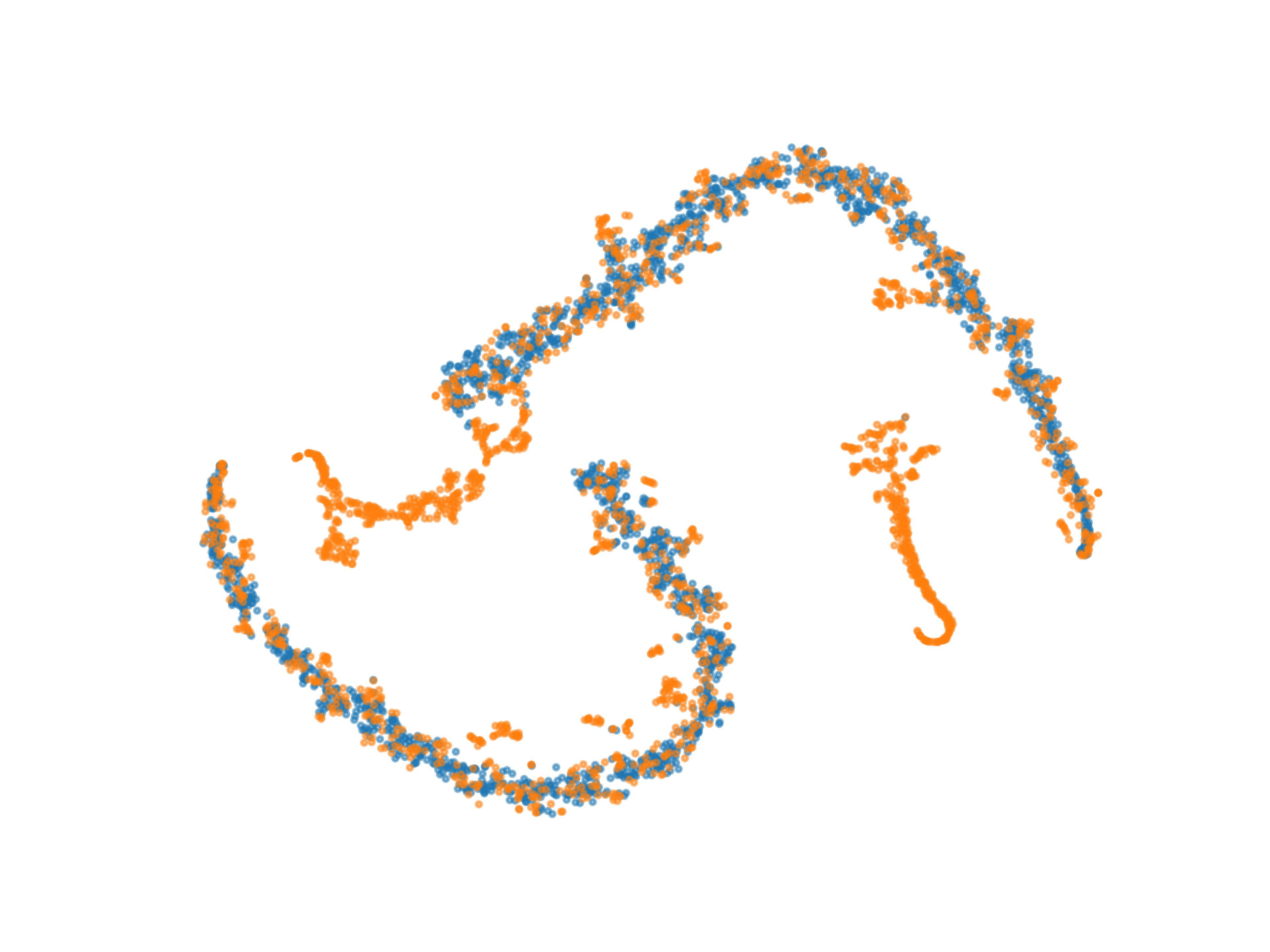}
        \captionsetup{font=small}
        \caption{SigCWGAN}
    \end{subfigure}%

    \caption{t-SNE visualizations. Top row: t-SNE plot where blue, green, and red marks data of dynamics 0,1,2.  Bottom row: t-SNE plot of real data $R$ as blue dots with $F$ as orange dots. The Market-GAN uses the C-TimesBlock.}
    \label{fig:T-SNE_combined}
\end{figure*}

\section{Experiments}
\subsection{Benchmarks and Datasets}
We utilized the daily OHLC features of 29 stocks (excluding DOW due to insufficient data) from the DJI index, from January 2000 to June 2023. This is a large dataset that guarantees generalization. 
We compared our model with representative generative models for time-series generation tasks including TimeGAN, SigCWGAN, CGMMN, and RCGAN. In the experimental setup, $\mathbf{X}$ represents a continuous segment of the price feature possessing identical $d$ and $l$ attributes and a length of 30 while $D=3$ and $L=29$. $\mathbf{H}$ is the preceding segment of features relative to $\mathbf{X}$, also with a length of 30. \\
\noindent\textbf{Multi-expert Benchmarks.} As not all benchmarks are designed for contextual generation, we applied a multi-expert paradigm when training the benchmark models for fairness. We train and evaluate a respective model $M_{i,j}$ of each benchmark method with the sub-dataset $R_{i,j}=\{(\mathbf{X}, \mathbf{H},d,l)|d=i,l=j\}$, with $\mathbf{H}$ as the conditional input for the benchmarks.
We show the result as the average score of all the experts of a model. 
\subsection{Ablation Study}
\noindent\textbf{Data Transformation Layer.} We evaluate the benchmarks with the application of the data transformation layer and inverse transformation in the pipeline. The results are included in the Table \ref{table:Data Usability Results} and the brackets of Table \ref{table:Generation Alignment Results}, \ref{table:Generation Fedility Results and Market Facts Result}.

\noindent\textbf{C-TimesBlock.} We compare the performance of Market-GAN using RNN as building blocks in networks versus using C-TimesBlock (CTB).

\subsection{Quantitative Results}
\label{sec: Quantitative Results}
\begin{table}[h!]
\centering
\small
\begin{tabular}{c|c|c}
\hline
Model   & $L_{d}$ & $L_{l}$\\
\hline
Real Data  & 0.055  & 0.023  \\
\hline
SigCWGAN  & 2.758(2.370)  & 0.594(0.468) \\
TimeGAN  & 2.847(1.839) & 0.533(0.490)\\
CGMMN  & 2.415(2.181) & 0.560(0.489) \\
RCGAN  & 3.276(2.279) & 0.574(0.492) \\
\hline
Market-GAN(CTB)  & \textbf{0.023}  & \textbf{0.100}\\
Market-GAN(RNN)  & 0.626  & 0.501 \\
\hline
\end{tabular}
\caption{Alignment Result (results in the bracket are applied with the data transformation layer).}
\label{table:Generation Alignment Results}
 \end{table}

\noindent\textbf{Context Alignment.} As shown in Table \ref{table:Generation Alignment Results}, Market-GAN with C-TimesBlock outperforms baseline experts on $L_{d}$ and $L_{l}$ significantly. In addition, CTB outperforms RNN when used in Market-GAN. A $L_{d}$ which is lower than that of the real test data indicates Market-GAN with CTB has learned the latent semantic of market dynamics based on the coarse semantic $d$ label by the market dynamics model. With the autoencoder mining factors, Market-GAN successfully aligns the dynamics semantic to $F$. While we observe a steady improvement of both $L_{d}$ and $L_{l}$ on baseline methods when applied with the data transformation layer, the alignment result is still unacceptable. According to the result, we can confirm that the Market-GAN architecture, C-TimeBlock, and the data transformation layer contribute to a better context alignment.

\setlength{\tabcolsep}{2pt}
\begin{table}[h!]
\centering
\footnotesize
\begin{tabular}{c|c|c}
\hline
Model  & $L_{D}$ &   $L_{f}$ \\
\hline
SigCWGAN & $9.28\pm 6.11 (7.54 \pm 3.72)$ & 0.211(0)\\
TimeGAN  &  $5.67\mathbf{\pm}  2.53(6.52 \pm 5.25)$ & 0.150(0) \\
CGMMN  & $12.43 \pm 7.37 (\textbf{4.12} \pm \textbf{2.30})$  & 0.501(0)\\
RCGAN  & $6.01 \pm 2.85 (7.07 \pm 6.68)$ & 0.247(0) \\
\hline
Market-GAN(CTB)   & $8.05 \pm 5.60 $ &\textbf{0}\\
Market-GAN(RNN)   & $11.85 \pm 9.89 $  &\textbf{0}\\
\hline
\end{tabular}
\caption{Fidelity result $L_{D}$ (results in the bracket are with the data transformation layer) and market facts result $L_{f}$.}
\label{table:Generation Fedility Results and Market Facts Result}
 \end{table}

\noindent\textbf{Fidelity.} 
\label{paragrah:Fidelity}
We train 87 discriminators with 3 dynamics and 29 stock tickers whose result is shown in the $L_{D}$ column of Table \ref{table:Generation Fedility Results and Market Facts Result}. Market-GAN with CTB got the 3rd lowest $L_{D}$ among the 5 compared methods. For our contextual generation task, alignment to control semantics and replicate $R$ can be adversary objectives. We observe that while the data transformation layer reduces $L_{d}$ and $L_{l}$, applying it also increases the $L_{D}$ of TimeGAN and RCGAN, whose $L_{D}$ is lower than that of Market-GAN. This phenomenon replicates the FID, CLIP score trade-off which has been observed in text-to-image generation \cite{chang2023muse}. While our generation goal is beyond letting $R=F$, minimizing $L_{D}$ isn't the ultimate goal. Mark-GAN outperforms the benchmark methods by improving the context alignment while maintaining decent fidelity. Given these considerations, it is acceptable that Market-GAN doesn't achieve the minimum $L_{D}$.

\setlength{\tabcolsep}{3pt}
\begin{table*}[h]
\centering
\footnotesize
\begin{tabular}{c|c|c|c|c}
\hline
Training Set Data & TimesNet & TCN & LSTM & GRU  \\
\hline
Only Real Data  &  $1.76 \pm 0.71 $ &  $ 6.63 \pm 0.76  $ &  $ 20.9\pm 20.9 $  &  $ 17.2 \pm 17.9  $     \\
 \hline
 SigCWGAN w/o TF &  $2.71\pm 8.07 $  &   $ 15.0 \pm 26.3   $ &  $19.1\pm 34.8 $  &  $ 17.6  \pm 33.7  $  \\
 SigCWGAN w TF &  $ 5.64\pm 12.7 $  &  $18.1\pm 22.8 $ &  $ 24.0 \pm 25.0$  &  
$20.7  \pm 23.7  $  \\
 TimeGAN w/o TF & $1.73\pm0.72$ & $6.59\pm 0.70 $  &  $19.5\pm 20.1$  &  $ 16.6 \pm 17.5 $  \\
 TimeGAN w TF & $ 1.36\pm0.48$ & $6.52 \pm 0.72$ &  $ 19.1 \pm 19.8  $  &  $ 14.8 \pm 16.3  $  \\
 CGMMN w/o TF & $1.74\pm4.98$ & $8.92\pm12.3$ &  $ 11.1 \pm 19.4 $  &  $ 8.74\pm 16.5  $ \\
 CGMMN w TF & $1.38\pm0.50$ & $ 6.53\pm 0.73 $ &  $ 19.1 \pm 19.8 $  &  $ 14.8 \pm 16.3 $  \\
 RCGAN w/o TF & $1.68\pm0.67$ & $6.64\pm0.83$ &  $ 16.6 \pm 18.1 $  &  $ 13.9  \pm 15.5$  \\
 RCGAN w TF & $1.40\pm0.55$ & $6.70\pm1.91$ &  $ 19.2 \pm 19.7 $  &  $ 15.0 \pm  16.2 $ \\
 \hline
  Market-GAN(CTB) & $1.26\pm0.75$  & $6.48 \pm 0.73$ & $12.9 \pm 16.7 $  &  $ 11.1 \pm 14.2 $  \\
 Market-GAN(RNN) &  $\textbf{1.11} \mathbf{\pm} \textbf{0.66} $  
& $\textbf{6.39} \mathbf{\pm} \textbf{0.92} $  
& $\textbf{9.89} \mathbf{\pm} \textbf{14.9} $  
& $\textbf{8.41} \mathbf{\pm} \textbf{12.4}$ \\
\hline
\end{tabular}
\caption{Prediction SMAPE loss on the test set using generated data of different models to augment the training set. TF is the abbreviation of the data transformation layer.}
\label{table:Data Usability Results}
\end{table*}

\noindent\textbf{Market Facts.} As shown in the $L_{f}$ column of Table \ref{table:Generation Fedility Results and Market Facts Result}, Market-GAN guarantees 0 violation of the facts on data with its constrained generation pipeline while baseline methods don't capture the fact without the data transformation layer.

\noindent\textbf{Data Usability.} We train 87 predictors with 3 dynamics and 29 stock tickers, whose results are shown in Table \ref{table:Data Usability Results}. While using generated data from benchmark methods results in a prediction loss that is comparable to or worse than using only real data, the generated data from Market-GAN consistently lowers the prediction loss on the test set. Market-GAN outperforms the benchmarks across all forecasting models, getting the lowest prediction loss on the test set, showcasing its effectiveness and reliability when applied to the down-stream prediction task.

\subsection{Qualitative Results}
\label{sec:Qualitative Results}
With research on limitations of the FID \cite{parmar2022aliased} and CLIP score \cite{radford2021learning}, recent works \cite{saharia2022photorealistic} rely on human expert ratings instead of quantitative metric. However, due to the complexity of financial data, rating the fidelity and context alignment of Market-GAN with human experts is not plausible. As a counterpart, we visualize data with t-SNE plots.\\
\noindent\textbf{Context Alignment.}
\label{paragrah:Context Aligment}
We plot the t-SNE plot of $F$ where the data of different dynamics is marked respectively as illustrated in the top row of Fig \ref{fig:T-SNE_combined} as a qualitative evaluation of the $L_{d}$. The t-SNE plot of $R$, as shown in Fig \ref{fig:MDM_result}(b), shows a separated cluster of data from dynamics $0$ and $2$ while data from dynamics $1$ is spread among the two clusters. While this pattern is replicated by the benchmarks to a different extent, we observe that $F$ of Market-GAN has three distinct clusters, indicating that it discovers a more disentangled representation of $d$, resolving mode collapse and leading to a more diverse generated $F$, corresponding to the numerical result that Market-GAN has a lower $L_{d}$ than real data $R$.\\
\noindent\textbf{Comparison of R and F.} We visualize the $F$ with $R$ with t-SNE plots as shown in the bottom row of Fig \ref{fig:T-SNE_combined}. This graphical representation provides further insights into the sources of $L_{D}$ loss. i) While CGMMN and RCGAN have high $L_{D}$ values (12.43 and 9.28) their t-SNE plot shows there are $F$ clusters that are separated from the $R$ clusters as outliers; ii) TimeGAN and RCGAN have low $L_{D}$ values (5.67 and 6.01) suggesting successful reconstruction of $R$. Thus, their $F$ overlaps with $R$ in visualization; iii) Market-GAN, with a moderate $L_{D}$ (8.05), does not contain significant outliers $F$ clusters from $R$. Instead, the points in $F$ that are not overlapping with $R$ are still adjacent to the cluster. With the discussion of a low than-real $L_{D}$ of Market-GAN, these non-overlapping $F$ points could be markets with extreme dynamics of $d$ that does not present in the historical data. With a strong ability to control generation with context, Market-GAN is able to simulate extreme markets \cite{orlowski2012financial} which could be useful in downstream financial applications. 

\section{Conclusion}
In this research, we present a financial simulator Market-GAN with the Contextual Market Dataset. With the innovative hybrid model devised for the contextual generation of financial data, Market-GAN surpasses existing methods in generating context alignment data which can improve downstream task performance while maintaining fidelity.
Our model offers extensive potential for enhancements and broadened applications. One such possibility is the integration of more fundamental factors beyond stock tickers and accommodating financial data of varied structures and scales. 
In essence, Market-GAN not only sets a new benchmark in the generation of financial data but also heralds exciting prospects for the evolution of this model, paving the way for the next generation of financial simulation.

\section*{Acknowledgments}
This project is supported by the National Research Foundation, Singapore under its Industry Alignment Fund – Pre-positioning (IAF-PP) Funding Initiative. Any opinions, findings and conclusions or recommendations expressed in this material are those of the author(s) and do not reflect the views of National Research Foundation, Singapore. 

\bibliography{aaai24}

\end{document}